\documentclass{article}

\usepackage{arxiv}
\usepackage{amsmath}
\usepackage[utf8]{inputenc} 
\usepackage[T1]{fontenc}    
\usepackage{hyperref}       
\usepackage{url}            
\usepackage{booktabs}       
\usepackage{amsfonts}       
\usepackage{nicefrac}       
\usepackage{microtype}      
\usepackage{lipsum}
\usepackage{graphicx}
\graphicspath{ {./images/} }
\usepackage{multirow}

\title{Visual Prompt Based Reasoning for Offroad
Mapping and Navigation using Multimodal LLMs}

\author{
 Abdelmoamen Nasser \\
  Robotics\\
  Khalifa University\\ 
  Abu Dhabi, United Arab Emirates\\
  \texttt{abdelmoamen.nasser@ku.ac.ae} \\
   \And
 Yousef Baba'a \\
  Computer Science\\
  Khalifa University\\ 
  Abu Dhabi, United Arab Emirates\\
  \texttt{yousef.babaa@ku.ac.ae} \\
  \And
 Murad Mebrahtu \\
  Computer Science\\
  Khalifa University\\ 
  Abu Dhabi, United Arab Emirates\\
  \texttt{murad.mebrahtu@ku.ac.ae} \\
  \And
 Nadya Abdel Madjid\\
  Computer Science\\
  Khalifa University\\ 
  Abu Dhabi, United Arab Emirates\\
  \texttt{nadya.madjid@ku.ac.ae} \\
  \And
 Jorge Dias \\
  Computer and Information Engineering\\
  Khalifa University\\ 
  Abu Dhabi, United Arab Emirates\\
  \texttt{jorge.dias@ku.ac.ae} \\
  \And
 Majid Khonji \\
  Computer Science\\
  Khalifa University\\ 
  Abu Dhabi, United Arab Emirates\\
  \texttt{majid.khonji@ku.ac.ae} \\
}

\begin{document}
\maketitle
\begin{abstract}
Traditional approaches to off-road autonomy rely on separate models for terrain classification, height estimation, and quantifying slip or slope conditions. Utilizing several models requires training each component separately, having task specific datasets, and fine-tuning. In this work, we present a zero-shot approach leveraging SAM2 for environment segmentation and a vision-language model (VLM) to reason about drivable areas. Our approach involves passing to the VLM both the original image and the segmented image annotated with numeric labels for each mask. The VLM is then prompted to identify which regions, represented by these numeric labels, are drivable. Combined with planning and control modules, this unified framework eliminates the need for explicit terrain-specific models and relies instead on the inherent reasoning capabilities of the VLM. Our approach surpasses state-of-the-art trainable models on high resolution segmentation datasets and enables full stack navigation in our Isaac Sim offroad environment.
\end{abstract}


\section{Introduction}
Alongside sustained progress in urban autonomy, an increasing body of research has begun to address the challenging problem of off-road navigation \cite{Maturana2017RealTimeSM, Shaban2021SemanticTC}. The intrinsic complexity of off-road navigation arises from the absence of high-fidelity maps, the lack of precise localization, and the diversity of terrains. However, despite these challenges, off-road autonomy is becoming an increasing priority, as most of the Earth’s surface is off-road and reliable navigation remains an open problem, with critical applications in agriculture \cite{kabir_terrain_2025}, planetary exploration \cite{chung_pixel_2024}, and search-and-rescue missions \cite{basri_hybrid_2022}.

Since reliable navigation depends on understanding which regions of the environment are passable, some works focus on traversability estimation, evaluating whether a terrain can be safely crossed. One strategy is to train models for separate tasks, including terrain classification \cite{kabir_terrain_2025}, elevation estimation via monocular elevation maps \cite{chung_pixel_2024}, and slip prediction on unpaved terrains \cite{basri_hybrid_2022}, the latter ensuring more precise control and stability. Another strategy is to integrate these factors into a unified framework, such as TerrainNet \cite{meng2023terrainnet}, which fuses multi-view RGB inputs with stereo depth in a BEV representation to capture both traversable ground surfaces and elevated structures. Another line of work addresses open-trail detection, where the goal is to identify continuous, visually discernible paths that can guide navigation in unstructured environments \cite{hassan_terrainsense_2024,hassan_pathformer_2024}. In this context, transformer-based models have been explored \cite{Jain2022OneFormerOT,9878483}, with PathFormer \cite{hassan_pathformer_2024} standing out as a transformer-based framework that leverages multi-scale deformable attention to generate free-space maps and predict continuous trails under dynamic terrain conditions. In addition to trail-based methods, drivable-area detection has also been explored \cite{wu_drivable_2024}, with approaches that classify terrain patches as traversable or not. While these approaches demonstrate progress in terrain perception, due to the diversity of off-road environments, the training corpora should be of a broad coverage, and gaps in coverage can restrict a model’s ability to generalize. 

A more flexible alternative, which may generalize better to unfamiliar terrains, is visual grounding --- the use of natural-language supervision to interactively localize visual regions. In particular, Referring Expression Segmentation (RES) \cite{towardsRES_survey_2024} aims to generate pixel-level segmentations corresponding to a natural-language query. These queries can include instructions such as “where is the drivable area,” “which area is easiest to traverse,” or “which trail should the vehicle take.” Approaches often rely on CLIP-based encoders for zero-shot grounding \cite{multimodalRES_survey_2025, clipseg_2021}, demonstrating how pretrained vision-language alignment can be extended to segmentation. Trainable variants such as LAVT \cite{lavt_2021}, LISA \cite{lisa_2023}, and GRES \cite{gres_2024} further enhance RES by integrating segmentation backbones and cross-modal transformers, but still predominantly focus on object-level references rather than higher-level scene reasoning.

To improve segmentation quality in open-world settings, recent work has combined language-conditioned prompting \cite{sam4mllm_2024}, open-vocabulary detection \cite{groundedsam_2024}, and cross-modal attention \cite{vlsam} to modulate SAM’s mask selection in response to natural-language input. However, while SAM excels at spatial precision, it remains unaware of deep semantics \cite{espinosa2024samanticsexploringsambackbone}, which may limit its performance in scenarios where understanding depends not only on identifying regions, but also on judging the degree of traversability. Unlike closed-world tasks such as segmenting cars or pedestrians, determining whether a rocky slope is crossable requires multi-step reasoning. This has motivated researchers to explore VLMs beyond simple matching capabilities, but rather toward contextual reasoning \cite{Xu2023DriveGPT4IE, 10495699}.

In regards to contextual reasoning, recent work has explored the use of VLMs mostly in urban contexts in the tasks of segmentation and detection \cite{leonardis_3d_2025, liu_vlpd_2023}, as well as scene understanding and spatial reasoning using multimodal-to-text formulations \cite{choudhary2023talk2bev}. Representative systems such as Talk2BEV \cite{choudhary2023talk2bev} and DriveVLM \cite{xiaoyu2024} leverage BEV representations or map priors, and have primarily been developed for structured, map-rich environments. Expanding beyond perception-focused pipelines, an emerging line of work explores Vision-Language-Action (VLA) models \cite{robonurse_2024, chen2023openvla}, which aim to translate natural-language instructions into goal-directed behaviors by tightly integrating perception, reasoning, and control. These models are particularly appealing for off-road autonomy due to their ability to handle contextual ambiguity, adapt to unseen scenarios, and perform multi-step reasoning grounded in language and sensory inputs. VLA agents are typically trained to interpret high-level tasks in context, condition their behavior on multimodal observations, and execute structured action plans through sequential decision-making. However, existing VLA systems have not yet been applied to outdoor or navigation-centric domains. For instance, RoboNurse \cite{robonurse_2024} operates in clinical settings to interpret medical instructions, while OpenVLA \cite{chen2023openvla} is designed for general-purpose robotics via visual-language feedback loops. Despite their focus on manipulation and assistive robotics, these architectures offer promising foundations for extending reasoning and control capabilities to terrain-aware off-road navigation.

Thus, the existing off-road solutions either require extensive labeled data to generalize across diverse terrains, or they rely on general-purpose language models without explicit mechanisms for grounding spatial semantics in complex environments. To address this gap, we propose a unified, zero-shot framework that integrates SAM-based segmentation with VLM-based reasoning to identify drivable regions. Our method overlays numeric labels on segmented regions and prompts the VLM to infer which regions are traversable based on both visual input and linguistic instructions. Coupled with lightweight planning and control modules, the framework enables high-level semantic understanding to be translated into actionable navigation behavior. We benchmark the perception module on standard off-road datasets and evaluate the complete system through goal reachability tests in a simulated environment. We claim the following contributions:
\begin{itemize}
    \item A unified zero-shot navigation framework that couples SAM-based segmentation with VLM-based reasoning to identify drivable regions, integrated with lightweight planning and control for end-to-end goal-directed behavior.
    \item A benchmark for off-road navigation that evaluates end-to-end goal reachability from raw sensory input to control. This benchmark enables future work to assess the full decision-making loop in unstructured environments in addition to segmentation labels within the simulated environment.
\end{itemize}

The rest of the paper is structured as follows: Section~\ref{sec:methodology} presents the proposed framework, including segmentation, reasoning, planning and control components. Section~\ref{sec:setup} describes the experimental setup and benchmark design. Section~\ref{sec:experiments} reports quantitative and qualitative results. Finally, Section~\ref{sec:conclusion} concludes the paper and outlines future directions.
\begin{figure*}[t]
    \centering
    \includegraphics[width=\textwidth]{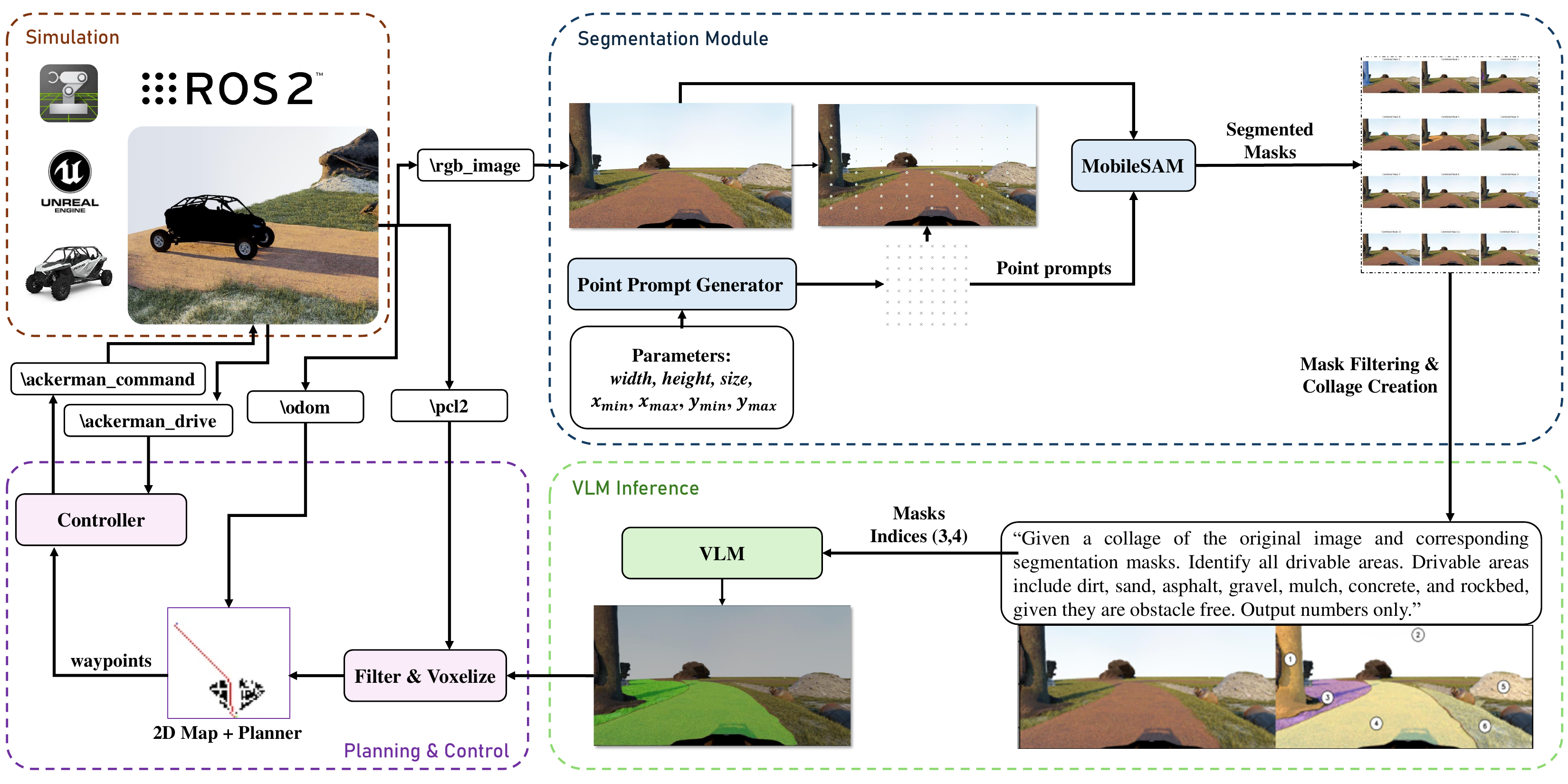}
    \caption{Overview of the off-road framework: (i) Simulation Module using NVIDIA Isaac Sim, (ii) Segmentation Module with SAM2 to for mask generation and tracking, (iii) VLM Inference Module to detect drivable area based on segmented masks and textual prompts, (iv) Mapping, planning, and control for environment mapping, path generation, and path following and velocity tracking respectively.}
    \label{fig:1}
\end{figure*}
\section{Methodology}
\label{sec:methodology}

The proposed off-road framework is detailed in Fig. \ref{fig:1}, comprising four main blocks:

\begin{enumerate}
\item Segmentation Module: First, a set of prompt points is generated, which are subsequently utilized by SAM2 \cite{SAM2} to generate segmentation masks. The generated masks are then refined and overlaid on the original image, producing an annotated collage. The same module is used for tracking the drivable area after VLM inference.
\item VLM Inference Module: This module combines the annotated image with the original image to create a side-by-side collage. Along with a textual prompt specifying drivable area classes, such as dirt, asphalt, and gravel, it queries the VLM to guide the model in outputting only the relevant indices of obstacle-free regions. The final output of this module is a binary mask indicating the drivable area.
\item Mapping, Planning and Control: These modules, respectively, processes the binary mask containing drivable areas and convert them into a 2D grid representation of the environment. A global planner is then used to compute the shortest path on this dynamically updated grid while a local planner generates a kinematically feasible trajectory. The control module employs a Stanley controller \cite{hoffmann_autonomous_2007} for lateral control and a PID controller is used to regulate velocity, with outputs for steering and acceleration.
\end{enumerate}  

\subsection{Segmentation Module}
The segmentation phase starts with generating a series of point prompts within the specified boundaries. The width and height are acquired from the image dimensions, while the minimum and maximum boundaries are set to control the distribution of the points over the image dimensions. This array is used to prompt SAM2, which takes a point prompt to identify the object at the coordinate of the point. Looping over the grid yields faster performance than the built in mask generation function of SAM2, making it more suitable for real time applications. The masks are then filtered to eliminate unnecessary masks by combining masks that meet a certain threshold, which is formulated as:
Let \( P = \{p_1, p_2, \ldots, p_n\} \) represent input prompts. For each \( p_k \), a binary mask \( M_k \) is generated and resized to match the image \( I \). Masks are combined iteratively based on the Intersection over Union (IoU):
\begin{equation}
\text{IoU}(M_i, M_j) = \frac{|M_i \cap M_j|}{|M_i \cup M_j|} \geq \tau_{\text{IoU}},
\end{equation}

with \( \lor \)-based merging. To eliminate smaller open spaces which could not be drivable, masks with area:

\begin{equation}
A(M) = \sum_{(x, y)} M(x, y) \geq \tau_{\text{area}},
\end{equation}

are retained where \( x, y \) are pixel coordinates. This filtering stage results in a final set of masks \( \mathcal{M} = \{M_1', M_2', \ldots, M_{n-1}', M_n' \) \}. Each mask $M_i \in \mathcal{M}$ is assigned a number $i$ matching the mask's index and placed at the center of mass $c_i$ of the mask $M_i$. The center of mass $c_i$ is computed as:

\begin{equation}
\mathbf{c}_i = \left( \frac{\sum x \cdot M_i'}{\sum M_i'}, \frac{\sum y \cdot M_i'}{\sum M_i'} \right).
\end{equation}

Each mask $M_i \in \mathcal{M}$  displayed with transparency and overlaid on \( I \), resulting in the annotated image \( I_{\text{annotated}} \). Furthermore, SAM2's memory encoder is utilized for mask tracking in the upcoming frames. This minimizes the amount of times the VLM needs to be prompted since we only prompt when the track of the drivable area is lost. To limit the computational requirement of tracking, the memory encoder queue is cleared every 20 frames.

\subsection{VLM Inference Module}
To achieve the desired results, the annotated image generated in the previous module is placed alongside the original image to create a side-by-side collage. This visual collage is used to prompt a VLM, which also receives a textual prompt to guide its processing, e.g.:

\begin{quote}
\textit{''Given a collage of original image and corresponding segmentation masks. Identify all drivable areas. Drivable areas include dirt, sand, asphalt, gravel, mulch, concrete, and rockbed given they are obstacle free. Output numbers only.''}
\end{quote}

The classes of drivable areas included in the prompt are acquired empirically from open-trail detection research. The numerical data retrieved from the VLM’s response is then used to filter the mask array which is aligned with the annotated numbers through indices. These numbers act as criteria for isolating the relevant elements within the mask array, ensuring a precise and targeted analysis. This approach combines visual and textual inputs to generate a binary mask indicating drivable areas. \\
\subsection{Mapping, Planning \& Control}
The proposed planning framework adopts a two-level structure consisting of a global planner and a local planner. The global planner computes a coarse path across the full environment, represented as a 12,000×12,000 map discretized into a 600×600 occupancy grid. After the segmentation mask is passed to the planning module, the point cloud is processed to mark drivable areas with 0s and all other areas with 1s. This is done through indexing since image and pointcloud coordinates are calibrated within the simulated version of the ZED camera. The grid is constructed by voxelizing the processed point cloud into discrete cells, each representing a region in the environment.This processed map is voxelized into grid cells that provide the basis for both global and local planning. The global planner generates a long-horizon path toward the goal, while the local planner refines this path in real time using the dynamically updated occupancy grid. By continuously adjusting to new segmentation and point cloud updates, the local planner ensures the path remains feasible and aligned with the vehicle’s motion constraints. This two-level design balances efficiency and adaptability, combining stable goal-directed navigation with responsive maneuvers in dynamic environments.\\
\begin{figure*}[t]
	\centering
	\includegraphics[width=\linewidth]{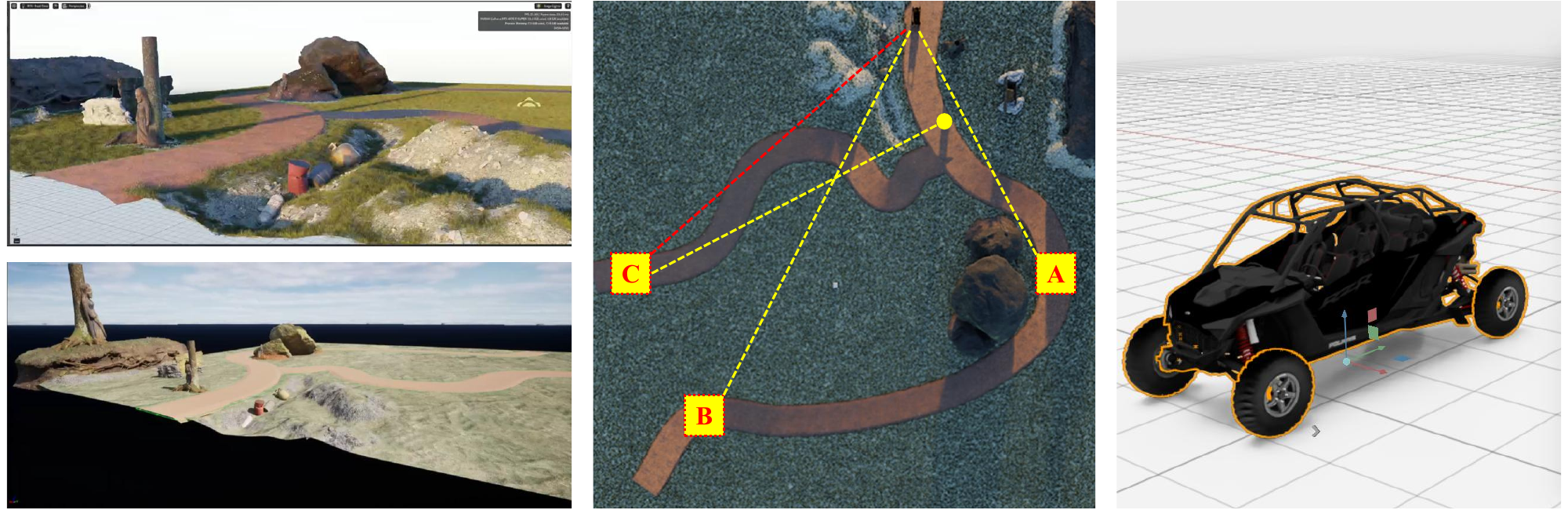}
	\caption{Simulation Setup: Bottom left: Simulation environment created in Unreal Engine. Top left: Simulation environment imported into Isaac Sim. Center: Top view of the open trail created using the spline creation tools, where the letters (A,B and C) indicate the goals in the reachability test. Right: Polaris RZR Sport 2022 imported using PhysX vehicle api.}
	\label{fig:2}
\end{figure*}
The voxelization process reduces the complexity of the environment by downsampling points from the point cloud into discrete grid cells. The voxel coordinates for each point $\mathbf{p} = [x, y]$ are computed as:

\begin{equation}
\mathbf{v} = \left\lfloor \frac{\mathbf{p}}{\mathbf{s}} \right\rfloor
\end{equation}

where $\mathbf{s} $ is the downsampling factor and$\lfloor \cdot \rfloor$ represents the floor operation to map continuous coordinates into discrete voxel indices. 
The voxelized points are mapped onto a 2D grid, where cells are marked as $1$ if occupied by obstacles and $0$ if free space. The grid serves as the input for the global and local planner.

The global planner is implemented using D* lite \cite{dlite}, which efficiently computes a shortest path while supporting incremental updates when the map changes. Unlike static search methods, D* Lite reuses previous search results, making it well-suited for dynamic environments where obstacles or free space may be updated.
The algorithm assigns each node $v$ two values, $g(v)$ and $rhs(v)$, both initialized to $\infty$ except for the goal node $g$, where $rhs(g) = 0$. A priority queue $U$ is initialized using a key function based on the minimum of $g(v)$ and $rhs(v)$ plus a heuristic $h(s,v)$ (Euclidean distance in our case).

At each step, the node with the smallest key is expanded, updating its cost and those of its predecessors. When the occupancy grid changes, only the affected nodes are updated, avoiding full recomputation.The algorithm terminates when $g(s) = rhs(s)$, and the path is reconstructed by following successors that minimize $g(v) + c(v,u)$.

The local planner is implemented using Hybrid A* \cite{Dolgov2008PracticalST} and operates within a moving window along the global path. As its immediate goal, it selects a waypoint from the global trajectory that lies within this local planning horizon. By incorporating continuous heading angles and the vehicle’s kinematic constraints (e.g., turning radius), Hybrid A* generates dynamically feasible trajectories rather than grid-constrained paths.

The local planner initializes with the vehicle’s current state and a target point from the global path within its planning window. The search space is defined on the occupancy grid, with nodes extended using motion primitives consistent with the vehicle model.At each step, the node with the lowest cost is expanded. The cost function combines the traveled distance, a heuristic to the local goal, and penalties for infeasible or sharp maneuvers. Successors are generated using feasible steering actions, ensuring compliance with the vehicle’s motion constraints.

The algorithm terminates once the local goal is reached, reconstructing a smooth trajectory that connects to the global path. This allows the vehicle to perform responsive, dynamically compliant maneuvers in real time while remaining consistent with the overall navigation objective.

For lateral control, Stanley aims to minimize the cross-track error (CTE) between the vehicle's position and the generated path, and the heading error between the vehicle's orientation and desired direction along the path. For longitudinal control, a PID controller is utilized to ensure velocity tracking. The outputs of these two parts are steering angle and acceleration respectively which are both limited to the mechanical constraints of the vehicle. The controller assumes the kinematic bicycle model and follows the lateral control law:

\thickmuskip=0.2\thickmuskip
\begin{equation}
\small
\delta(t) =
\begin{cases} 
\psi(t) + \arctan\left(\frac{k e(t)}{v(t)}\right) & \text{if } \left| \psi(t) + \arctan\left(\frac{k e(t)}{v(t)}\right) \right| < \delta_{\text{max}} \\ 
\delta_{\text{max}} & \text{if } \psi(t) + \arctan\left(\frac{k e(t)}{v(t)}\right) \geq \delta_{\text{max}} \\
-\delta_{\text{max}} & \text{if } \psi(t) + \arctan\left(\frac{k e(t)}{v(t)}\right) \leq -\delta_{\text{max}}
\end{cases}
\end{equation}

where $\psi(t)$ is the heading error, $e(t)$ is the cross track error, and $\delta_{max},\delta_{min}$ are the maximum steering angles.
The longitudinal control law is as follows:
\begin{equation}
\ddot{x}_{\text{des}} = K_P (\dot{x}_{\text{ref}} - \dot{x}) 
+ K_I \int_{0}^{t} (\dot{x}_{\text{ref}} - \dot{x}) \, dt 
+ K_D \frac{d(\dot{x}_{\text{ref}} - \dot{x})}{dt}
\end{equation}
where $\ddot{x}_{des}$ refers to the desired acceleration, $K_P , K_I , K_D$ refer to the proportional, integral and differential gains.

\section{Implementation Details}
\label{sec:setup}
\subsection{Simulation Setup}

The test environment, created from scratch, initially utilizes Unreal Engine tools for terrain sculpting. Procedural generation tools are then employed to create various environmental elements, such as rocks, grass, and sand, tailored to the specific ecosystem. Unreal Engine \footnote{Unreal Engine, \url{https://www.unrealengine.com/en-US}.} also provides a spline creation tool for open trail design. In addition, existing marketplace assets and tools such as Quixel \footnote{Quixel, \url{https://quixel.com/}.}  Bridge and Megascans were used to add cosmetic elements such as gas cylinders, stone wall, statues and huge rocks. NVIDIA Omniverse \footnote{NVIDIA Omniverse, \url{https://www.nvidia.com/en-us/omniverse/}.} provides an unreal engine connector which converts the unreal engine environment into a Universal Scene Description (USD) \footnote{OpenUSD, \url{https://openusd.org/release/intro.html}.} file for use inside Omniverse applications. Finally, using NVIDIA PhysX vehicle API, a 3D model of the Polaris RZR Sport vehicle \footnote{3D Models Online Collection, \url{https://3dmodels.org}.} is imported into the environment and equipped with a ZED X camera available in the simulator's sensor assets. The camera is placed on the top front rail on the roof of the vehicle. Snapshots of the created environment are shown in Fig. \ref{fig:2}. Note that all the data is published using ROS2 Humble. Segmentation only tests were conducted on an NVIDIA RTX 4070 16 GB GPU and an i9-9900K processor at 3.6 GHz with 32 GB of DDR5 RAM. Tests that require running simulation and models simultaneously for real-time evaluation were conducted on a similar setup with NVIDIA RTX 3090 24 GB GPU.

\subsection{Dataset Collection}
\begin{figure*}[t]
	\centering
	\includegraphics[width=\linewidth]{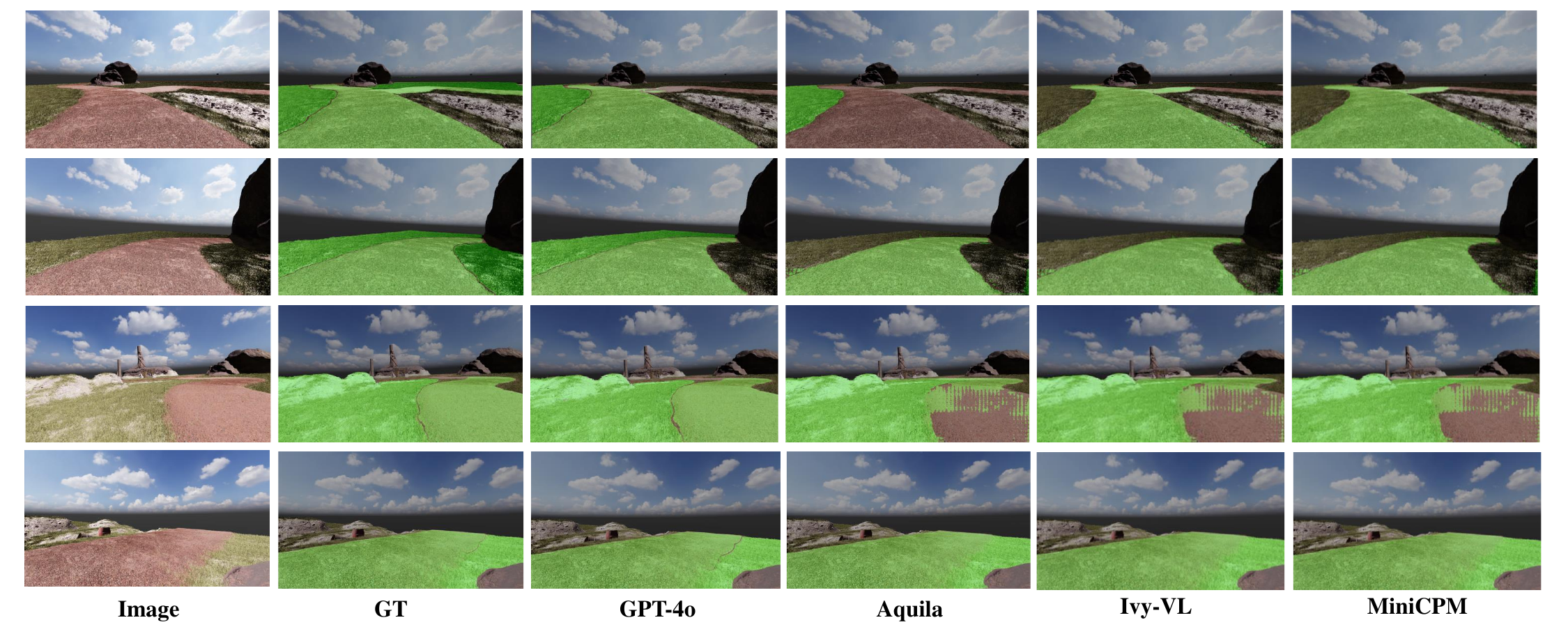}
	\caption{Samples of drivable areas detected by evaluated VLMs. GT refers to Ground Truth, while the remaining columns display the outputs of each model: ChatGPT-4o, Aquila, Ivy-VL and MiniCPM.}
	\label{fig:VLM_samples}
\end{figure*}

After setting up the simulation environment, the sensor equipped vehicle can be controlled via joystick with default controls from the vehicle API. Using the ROS2 capability of recording a rosbag, the topic publishing the RGB images is captured as the vehicle is driven around the environment. The recorded path ensures the capture of areas including grass, open trail, rocks and combinations of all the previous elements at different elevations. The rosbag is processed to extract individual frames resulting in a dataset of 2,957 RGB images. Samples of these images can be seen in Fig. \ref{fig:VLM_samples}.

\begin{figure*}[t]
	\centering
\includegraphics[width=\linewidth]{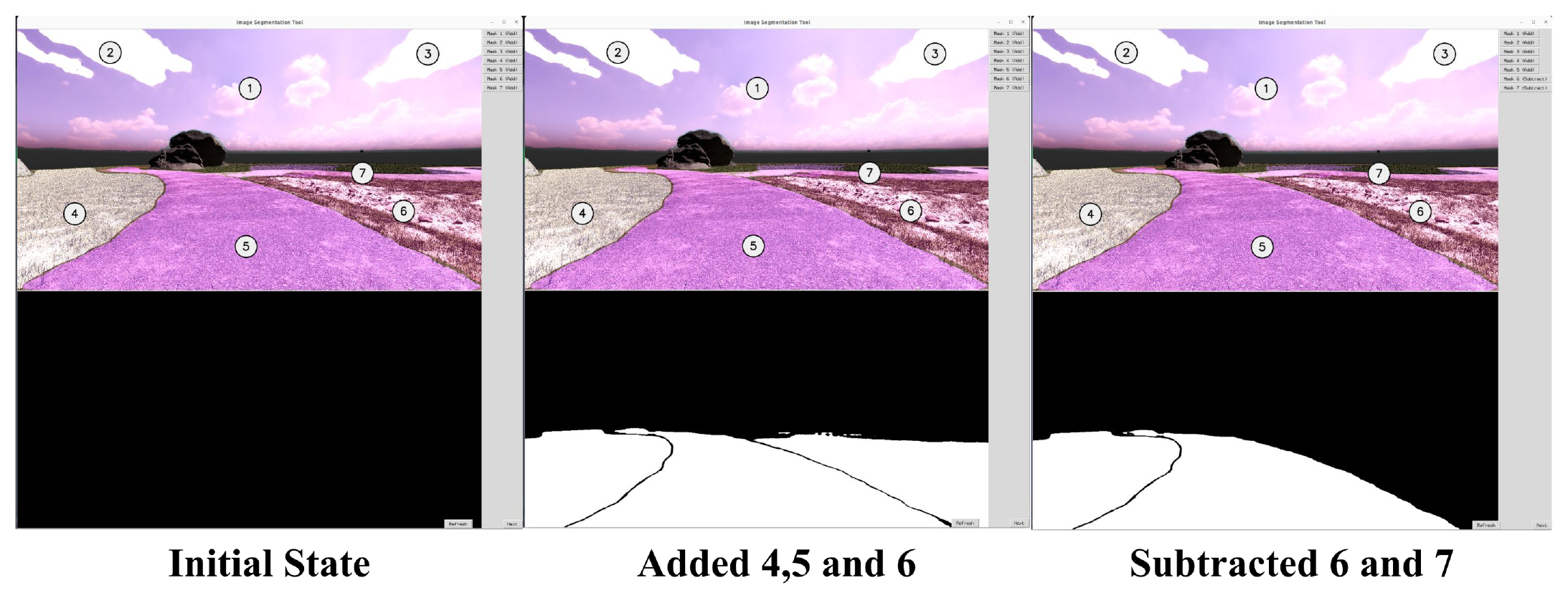}
	\caption{The pre-processing interface at three states from left to right. (i) The initial state. (ii) After addition of drivable masks 4, 5 and 6. (iii) After subtraction of masks 6 and 7.}
	\label{fig:ppgui}
\end{figure*}
\subsection{Pre-processing Interface}
The pre-processing stage utilized the existing segmentation module in the creation of ground truth labels for the collected dataset. The first stage is segmentation of the scene with the same parameters used in the complete pipeline in order to ensure consistency of segmentation vs ground truth. The images are processed in numerical order. An interface displays the annotated image to the user with check boxes corresponding to the displayed images. Below the original image is a blank (all black) image to represents an empty ground truth mask. To the side of these images, the user is provided with buttons corresponding to displayed annotation masks. The user can click on the mask once to add it to the ground truth, click twice to subtract it from the current ground truth, or click three times to reset the state of the chosen mask. This way, ground truth labels generated from the same segmentation technique used at inference time can be used for later on evaluation of the VLMs. Fig. \ref{fig:ppgui} provides a brief demonstration of the pre-processing interface.

\section{Experiments}
\label{sec:experiments}

This section presents the experiments conducted to evaluate the performance of the proposed off-road framework. The first set of experiments assesses the real-time capabilities of the framework, including its ability to successfully navigate to a given goal. The second set evaluates the performance of different Vision-Language Models (VLMs). The following VLMs were tested: GPT-5 Mini and ChatGPT-4o-latest\footnote{OpenAI, \textit{GPT-4 Technical Report}, 2023, \url{https://openai.com/research/gpt-4}.}, Aquila \cite{aquila}, Ivy-VL \cite{ivy-vl}, and MiniCPM \cite{minicpm}. All selected models are capable of running the entire off-road stack on a single GPU, ensuring that the framework remains lightweight and efficient. 

Due to computational constraints, running extensive experiments to identify the optimal VLM configuration for all settings was not feasible. Therefore, the evaluation of VLM performance was divided into two stages. In the first stage, various parameters were tested on a subset of samples, and a scoring function was used to identify the best configuration (see Section \ref{sec:vlm_scoring}). In the second stage, the performance of each VLM was quantified using its best configuration on the full dataset (see Section \ref{sec:vlm_quantative}).


\subsection{Real-time Applicability}
\label{sec:real_time}
\begin{figure*}[t]
	\centering
\includegraphics[width=\linewidth]{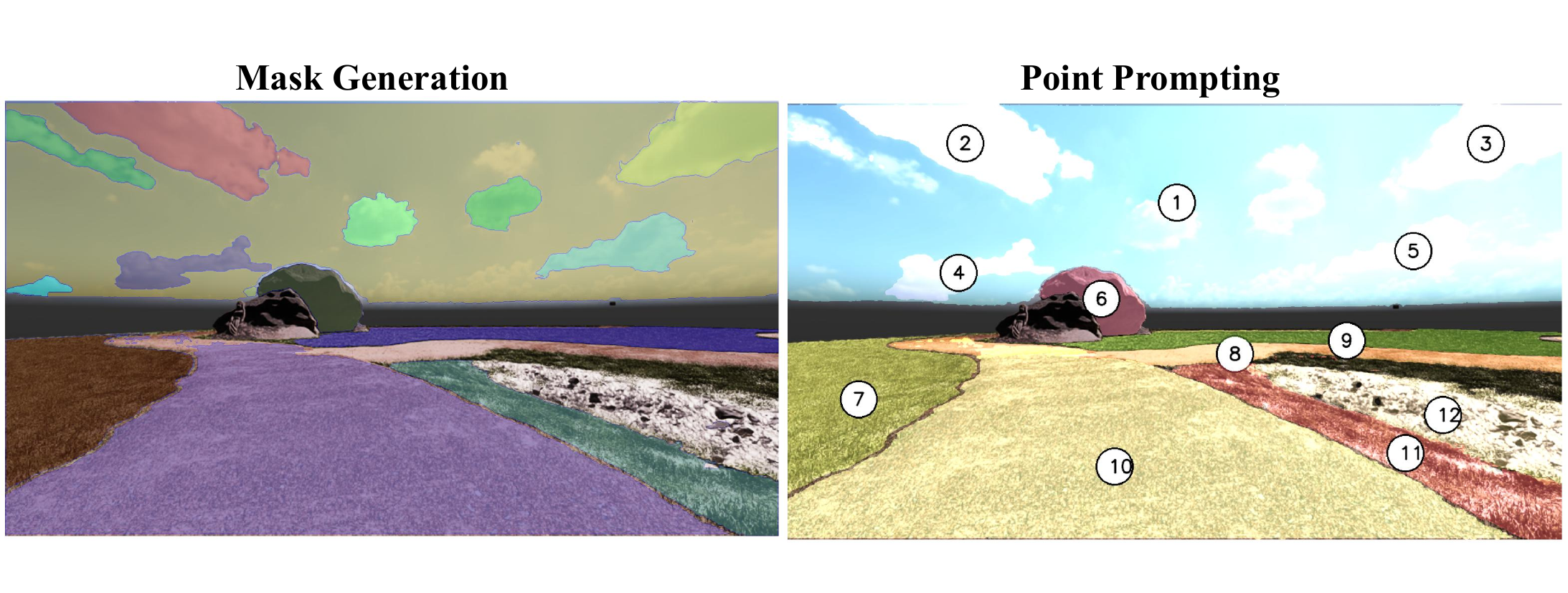}
	\caption{Qualitative comparison of Mask Generation vs Point Prompting on frame 88.}
	\label{fig:comppp}
\end{figure*}

SAM2's built in capability of automatically generating masks is compared with iterative point prompting for the four test images. The point prompting technique definitely yields faster results. The speed can vary based on the parameters passed where in this case both techniques are set at 64 point grid, a single point per forward pass, iou threshold at 0.5 and area threshold at 10000. The results observed in Table \ref{tab:ppgg} show that point prompting yields faster results. Fig. \ref{fig:comppp} shows qualitative samples of this test.
\begin{table}[h]
\centering
\begin{tabular}{lcc}
\toprule
Frame & Mask Generator   & Point Prompting  \\
\midrule
88  & 13.0s & 3.5s \\
500 & 12.8s  & 3.5s  \\
1500 & 12.5s  & 3.5s  \\
2000 & 12.7s  & 3.5s  \\
\bottomrule
\end{tabular}
\caption{Quantitative Comparison of Mask Generation vs Point Prompting.}
\label{tab:ppgg}
\end{table}

The reachability tested was conducted on three goals A,B, and C shown in Fig. \ref{fig:2}. The yellow lines indicate successful start and end coordinates. For goal C specifically, the initial start failed due to the simulated ditches. This exposes a weakness in the VLM detection as it does not identify these locations as obstacles. Therefore, from the starting point, the vehicle does not build up enough momentum to go over it. Alternatively, when the start was shifted to a farther location (yellow circle in Fig. \ref{fig:2}) the vehicle achieved a higher success rate than starting in the original position. Results are demonstrated in Table \ref{tab:rt} where every experiment was repeated 5 times to ensure robustness.
\begin{table*}[ht]
\centering
\begin{tabular}{|c|c|c|c|}
\toprule
\textbf{Start} & \textbf{Goal} & \textbf{Success rate} & \textbf{Avg. Time (s)} \\
\midrule
$S_1$ & A & 100\% & 72.56 \\
$S_1$ & B & 100\% & 163.08 \\
$S_2$ & C & 40\% & 114.97\\
\bottomrule
\end{tabular}
\caption{Reachability Test Results. $S_1$ indicates the original vehicle position. $S_2$ is the alternative position highlighted by the yellow circle in Fig. \ref{fig:2}.}
\label{tab:rt}
\end{table*}

\subsection{Evaluation of Different Prompt and Image Settings}
\label{sec:vlm_scoring}

\textbf{\textit{Overview:}}  This experiment was designed to evaluate how various parameters influence the VLMs' accuracy. The key variables examined were: (1) visual prompt format, (2) number of output masks' indices, and (3) linguistic prompt design.

To asses the impact of different \textit{visual prompt formats}, two configurations were designed. The first configuration("collage") is a composite collage that combines two images: the original image with its corresponding segmented and annotated counterpart. This side-by-side arrangement enables direct comparison between the raw input and its processed version, guiding the models to better interpret the scene. The second configuration(''annotated'') passes the segmented and annotated image to the VLM as the input without the original image. This design tests whether standalone processed outputs retain sufficient clarity and utility in the absence of comparative visual context.

The textual prompt can either instruct the VLM to output a \textit{single mask index} or allow \textit{multiple mask indices}. For this experiment both conditions were tested. The Single-Number Prompts (SNP) approach instructs the model to generate exactly one output mask, thereby imposing a fixed constraint on the quantity of generated results. On the other hand, the Multi-Number Prompts (MNP) omit explicit numerical constraints allowing the model to output as many masks as it deems necessary. This framework evaluates the trade-offs between rigid output control (SNP) and model-driven flexibility (MNP), displaying how prompt specificity can affect the consistency and adaptability of the models. \newpage
\begin{figure}[t]
	\centering
\includegraphics[width=\linewidth]{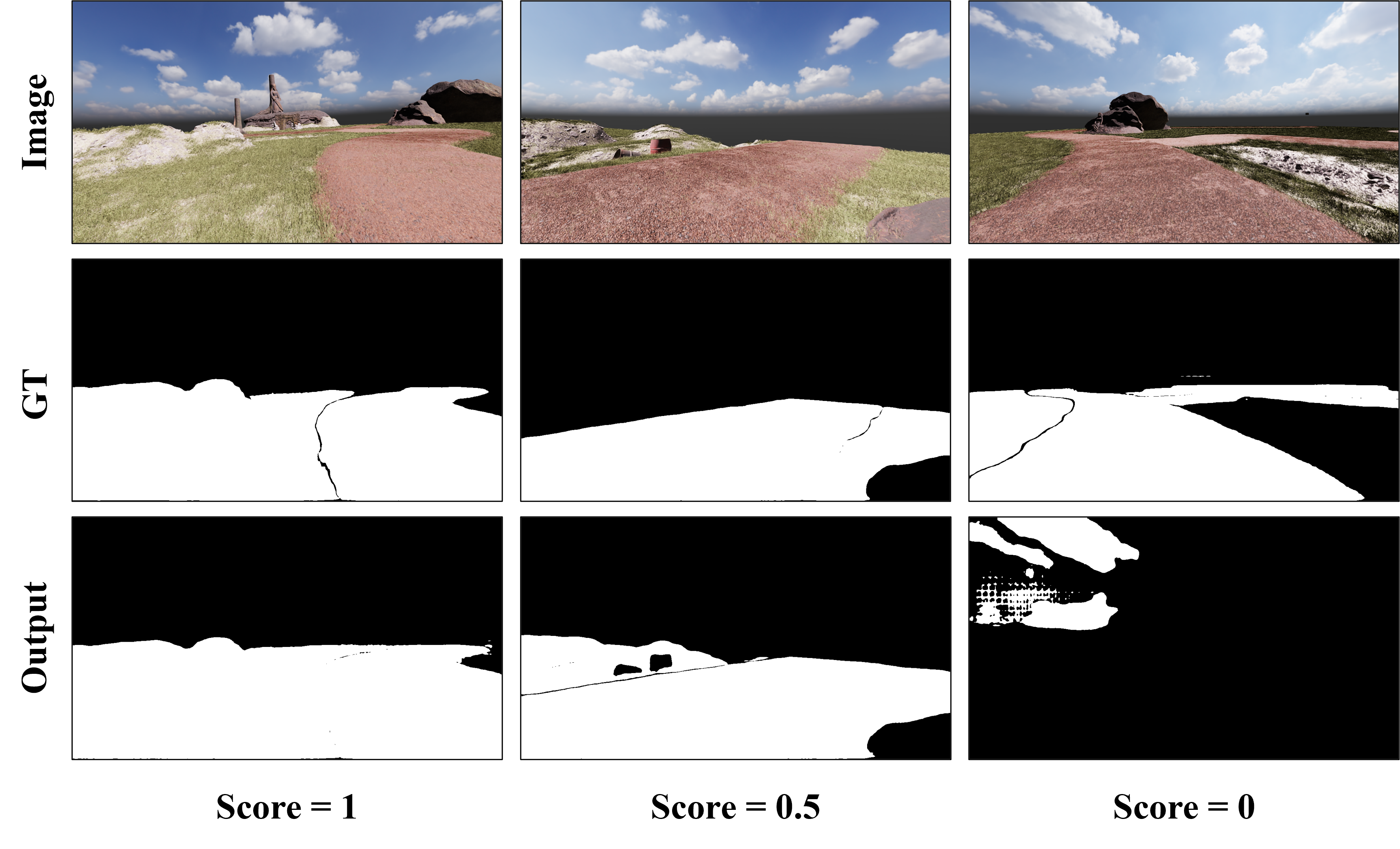}
	\caption{Examples from the scoring process: in the first column, a VLM selected mask indices that perfectly aligned with the Ground Truth (GT) drivable area, earning a score of 1; in the second column, a model selected indices corresponding to both a drivable area (an open trail) and a non-drivable area (a small hill of rocks), resulting in a score of 0.5; and in the third column, a model selected only mask indices corresponding to clouds, leading to a score of 0.}
	\label{fig:scoring_function_samples}
\end{figure}
To test different \textit{linguistic prompt designs}, we created three hierarchical prompt categories with varying amounts of contextual information provided:

\begin{itemize}
\item \textbf{Specific}: Direct requests for particular elements or terrain in the current scene. E.g., 
\begin{quote}\textit{
''Which mask number is the dirt path? Output number only.'', ''Which mask number is the dirt path and which mask number is the grass path? Output numbers only.''}
\end{quote}
\item \textbf{General}: Broad instructions to locate any drivable surfaces without specifying any details directly related to the environment. E.g.,
\begin{quote}\textit{
"Identify the number(s) on the mask(s) marking the drivable surface(s). Output number(s) only."}
\end{quote}
\item \textbf{Full Context}: 
A comprehensive yet concise task description that includes detailed instructions, contextual information, and environment-specific details. E.g., 
\begin{quote}\textit{
    ''You are a driving agent for an offroad car. Given a collage of original image and corresponding segmentation masks. Identify the drivable area. Drivable areas include dirt, sand, asphalt, gravel, mulch, concrete, patchy grass and rockbed. Output one mask number only.''}
\end{quote}
\end{itemize}

Considering all three variables and their variations, we created a total of 12 different combinations of visual and textual prompts. The evaluation was conducted on five images selected at regular intervals from the dataset (image numbers 88, 500, 1500, and 2000) to ensure diverse terrain features and segmentation challenges. Each combination of parameters was tested across all models using these five images.

\textbf{\textit{Metrics:}} To quantify the performance we used a scoring system based on visual inspection, where a score between 0 and 1 is assigned based on the following criteria: 
\begin{itemize}
    \item \textit{Score = 1}, if the VLM outputs all mask indices corresponding to drivable areas or selects a mask index that provides sufficient drivable space for the planner to navigate.
    \item \textit{Score = 0.5}, if the VLM outputs all mask indices corresponding to drivable areas or selects a mask index that provides sufficient drivable space for the planner to navigate, AND at least one mask index corresponding to non-drivable area. 
    \item \textit{Score = 0}, if the VLM outputs all the mask indices, OR only indices corresponding to non-drivable area.
\end{itemize}

The final score for each experiment was determined by summing the results from the four test images, with a maximum possible score of 5 points. Each configuration was tested five times on each model, and the average score was computed as the final result. Fig. \ref{fig:scoring_function_samples} presents examples from the scoring stage, with samples corresponding to each condition.
\begin{table*}[ht]
\centering
\resizebox{\textwidth}{!}{%
\renewcommand{\arraystretch}{1.1} 
\setlength{\tabcolsep}{8pt} 
\footnotesize 
\begin{tabular}{|p{5cm}|c|c|c|c|c|c|c|c|c|c|c|c|}
\hline
\multirow{3}{*}{\textbf{Model}} & \multicolumn{6}{|c|}{\textbf{Annotated}} & \multicolumn{6}{|c|}{\textbf{Collage}}\\ \cline{2-13}
 & \multicolumn{3}{|c|}{\textbf{MNP}} & \multicolumn{3}{|c|}{\textbf{SNP}} & \multicolumn{3}{|c|}{\textbf{MNP}}  & \multicolumn{3}{|c|}{\textbf{SNP}}\\ \cline{2-13}
 & \textbf{F} & \textbf{G} & \textbf{S} & \textbf{F} & \textbf{G} & \textbf{S} & \textbf{F} & \textbf{G} & \textbf{S} & \textbf{F} & \textbf{G} & \textbf{S} \\ \hline
\textbf{ChatGPT-4o-latest}         & 2.9            & 3.7         & 3.7        & 4.2         & 3.8           & 2.6           & 3           & 3           & 2.9         &\textbf{4.4}       & 4           & 3.4         \\ \hline
\textbf{Aquila-VL-2B}              & 0.6            & 1.5         & 3.2        & 2.8         & 3.4           & \textbf{4}    & 0.1         & 0.1         & 1.3         & 3.2               & 2           & 1.6         \\ \hline
\textbf{Ivy-VL (3B)}               & 0.6            & 1.1         & 1.7        & 3.8         & \textbf{4}    & 3             & 0.6         & 0.5         & 2.2         & 0.2               & 2.6         & 1.8         \\ \hline
\textbf{MiniCPM-V-2\_6-int4 (8B)}  & 1.1            & 2.1         & 3          & 2.9         & \textbf{4}    & 2.8           & 0.8         & 1.5         & 3.3         & 1.8               & 3.8         & 2.4         \\ \hline
\end{tabular}%
}
\caption{Average evaluation scores (0-5 scale) comparing vision-language models across visual prompt format, prompt specifications, and instruction types. Columns represent hierarchical experimental conditions: primary division by input format (Annotated/Collage), secondary by number of output masks' indices (MNP/SNP), tertiary by linguistic prompt design (F=Full Context, G=General, S=Specific). Bold values indicate maximum scores per model-condition combination.}
\label{tab:ql_results}
\end{table*}

\textbf{\textit{Results:}} Table \ref{tab:ql_results} presents the results of the scoring stage, highlighting variations in scores across different configurations. The visual prompt format impacts model performance, with annotated images improving overall accuracy for all models. For example, Ivy-VL (3B) achieved a +1.05 increase in average score under annotated conditions. Additionally, models with a larger number of parameters generally outperformed smaller ones, with ChatGPT-4o-latest achieving the highest average score (3.47), followed by MiniCPM-V-2\_6-int4 (8B) with an average score of 2.46. Despite Ivy-VL (3B) having one billion more parameters than Aquila-VL-2B, it scored slightly lower on average (by 0.14), with a mean score of 1.98.
SNPs produced better results across all models, particularly for those with fewer parameters. For instance, Aquila-VL-2B achieved a +1.7 higher mean score on SNPs compared to MNPs, suggesting that smaller models underperform when given more flexibility but perform well under constrained outputs. Finally, linguistic prompt design also played a crucial role in model performance. ChatGPT performed best with Full Context prompts under SNPs, reaching peak accuracy of 4.4 when using collages, demonstrating its ability to integrate detailed contextual instructions while adhering to strict output constraints. In contrast, models with fewer parameters exhibited the opposite trend. Aquila performed optimally with Specific SNP prompts (4.0/5 annotated), while Ivy-VL achieved its highest accuracy with General SNP prompts (4.0/5 annotated). These findings indicate that limiting contextual information when prompting smaller models leads to better performance.
\subsection{Performance Evaluation of VLMs in Simulation}
\label{sec:vlm_quantative}
Quantitative evaluation for each model was conducted using the best configuration, i.e., the prompt that achieved the highest average intersection over union (IoU) between the pipeline's output and the ground truth masks. Additionally, the VLM’s inference time per sample was recorded to assess model efficiency. This evaluation is conducted on the simulation set to choose the optimal model for real-time evaluation and testing on real world datasets. Table \ref{tab:vlm_performance} summarizes the results of the quantitative performance evaluation.

The evaluation of VLMs revealed notable trade-offs between accuracy and efficiency across different model sizes. GPT family models scored the highest IoU where GPT-5-Mini achieved the highest performance with an mIoU of 0.727 and 4o scoring 0.557. This came at the trade off of inference time where they took significantly more time to output a result.

Among models that were run locally, Ivy-VL offered a balance between performance and speed, obtaining an mIoU of 0.477 with an inference time of 0.90 seconds. Similarly, MiniCPM demonstrated faster inference at 0.54 seconds, but at the cost of reduced accuracy (mIoU 0.436).  The most balanced performance was observed with Aquila-VL-2B, which achieved a competitive mIoU of 0.515 alongside a relatively low inference time of 0.59 seconds. This suggests that Aquila-VL-2B offers the most effective trade-off between accuracy and efficiency, making it a strong candidate between the local models. 

Since the VLM is not frequently prompted, we use 4o as the main model. In cases where 4o fails to identify drivable areas, GPT-5-Mini is prompted as a contingency plan.

\begin{table*}[h]
\centering
\renewcommand{\arraystretch}{1.1} 
\setlength{\tabcolsep}{8pt} 
\footnotesize 
\begin{tabular}{p{3.0cm}|c|c|}
\toprule
\textbf{Model} & \textbf{mIoU} & \textbf{mIt (s)} \\ 
\midrule
GPT-5-Mini        &  \textbf{0.727} & 11.86 \\ 
ChatGPT-4o-latest        &  0.557 & 3.98 \\ 
Aquila-VL-2B             &  0.515 & 0.59 \\ 
Ivy-VL (3B)              &  0.477 & 0.90 \\ 
MiniCPM-V-2\_6-int4 (8B) &  0.436 & \textbf{0.54} \\ \bottomrule
\end{tabular}
\caption{Quantitative performance evaluation of VLMs on their optimal configuration. (mIt) stands for Mean Inference Time}
\label{tab:vlm_performance}
\end{table*}

\subsection{Evaluation on Real Off-road Datasets}

To assess the real world feasibility, the model was tested on the testing splits of ORFD, RUGD, and O2DTD datasets. For datasets with more terrain diversity (e.g. RUGD), the best evaluation result out of 5 is used due to the undeterminism in VLM predictions. Additionally, note that the VLM is prompted to identify open trail for ORFD and O2DTD but prompted for drivable area in RUGD. Results are shown in Table \ref{tab:qtd}. Testing splits and model results are following \cite{hassan_pathformer_2024}.
\begin{table}[h]
\centering
\begin{tabular}{lc}
\toprule
Model & Average\\
\midrule
Mask2Former \cite{Mask2Former}  & 0.7385\\
CLIPSeg \cite{CLIPSeg} & 0.6429\\
CGNet \cite{wu2020cgnet}& 0.5541\\
PSPNet \cite{zhao2017pspnet}&  0.7461\\
GroupViT \cite{xu2022groupvitsemanticsegmentationemerges}& 0.6899\\
OCRNet \cite{OCRNet}& 0.6119 \\
PathFormer \cite{hassan_pathformer_2024}& \textbf{0.7781}  \\
\midrule
Ours &  0.6905\\
\bottomrule
\end{tabular}
\caption{Quantitative performance evaluation (IoU) of proposed pipeline vs existing non-zero shot technique benchmarks averaged on the three datasets.}
\label{tab:qtd}
\end{table}

\begin{table}[h]
\centering
\begin{tabular}{lccc}
\toprule
Model & ORFD\cite{min2022orfddatasetbenchmarkoffroad} & RUGD\cite{RUGD2019IROS}  & O2DTD \\
\midrule
PathFormer \cite{hassan_pathformer_2024}& 0.7929  & 0.6447 & \textbf{0.8966} \\
Ours & \textbf{0.9141} & \textbf{0.8059} & 0.3516\\
\bottomrule
\end{tabular}
\caption{Dataset specific performance analysis.}
\label{tab:qtd}
\end{table}

The pipeline is generally on par with state-of-the-art models. However, it only surpasses them on datasets with high resolution images. This observation is further confirmed by the fact that it performs the worst on O2DTD\footnote{O2DTD, \url{https://avlab.io/datasets/offroad/}.} which does not show differentiating features of the drivable area, and by the fact that multi-scale \cite{hassan_pathformer_2024, Mask2Former, xu2022groupvitsemanticsegmentationemerges} or pyramid network \cite{zhao2017pspnet} based models are performing the best on average. This highlights a weakness of the current pipeline and the importance of multi-scale architectures in the off-road perception problem.

\section{Conclusion}
\label{sec:conclusion}

This paper introduces a novel zero-shot framework for off-road autonomous navigation that leverages SAM2 for segmentation and Vision Language Models (VLMs) for reasoning about drivable areas. By unifying segmentation and reasoning into a single pipeline, the proposed approach eliminates the need for traditional terrain-specific models and streamlines the perception process for off-road environments. The methodology, tested in a simulation environment created using Unreal Engine and NVIDIA Isaac Sim, demonstrates promising results with a simplified and effective pipeline for drivable area detection.

Results indicate that the VLM-driven approach closely approximates ground truth, showcasing its potential as a robust alternative to traditional methods. However, the generative and non-deterministic nature of some VLMs poses challenges in achieving consistent outcomes, highlighting areas for future research. Further exploration could involve fine-tuning VLMs, optimizing parameters like temperature and top-p constraints, and integrating additional modalities for enhanced reliability and generalization.

Overall, this work paves the way for leveraging modern segmentation and vision-language reasoning models to address the complex challenges of off-road autonomy, demonstrating a step forward in autonomous navigation systems. The future work will extend the framework towards: compatibility with lower resolution images and extension to Generalized RES, extension to dynamic environments and demabiguating semantics of objects, e.g., whether a specific path of grass is drivable or not based on density; and tuning the model's parameters (e.g., temperature and top-p) to ensure consistency, or alternatively fine-tuning VLMs to the specific task.

\bibliographystyle{unsrt}  
\bibliography{references}  

\end{document}